\documentclass[10pt,twocolumn,letterpaper]{article}

\usepackage{wacv}
\usepackage{times}
\usepackage{epsfig}
\usepackage{graphicx}
\usepackage{amsmath}
\usepackage{amssymb}

\usepackage{subfigure}
\usepackage{capt-of}

\def\wacvPaperID{907} 

\wacvfinalcopy 

\ifwacvfinal
\def\assignedStartPage{1} 
\fi

\ifwacvfinal
\usepackage[breaklinks=true,bookmarks=false]{hyperref}
\else
\usepackage[pagebackref=true,breaklinks=true,colorlinks,bookmarks=false]{hyperref}
\fi

\ifwacvfinal
\else
\fi

\begin{document}

\title{Color2Embed: Fast Exemplar-Based Image Colorization using Color Embeddings}

\author{Hengyuan Zhao\textsuperscript{\rm 1}\thanks{Co-first author. This work was done when Hengyuan Zhao was a research intern at Baidu VIS.} \qquad Wenhao Wu\textsuperscript{\rm 1}\footnotemark[1] \qquad Yihao Liu\textsuperscript{\rm 2}\footnotemark[1]  \qquad Dongliang He\textsuperscript{\rm 1} \\
\textsuperscript{\rm 1}Department of Computer Vision Technology (VIS), Baidu Inc.\\
\textsuperscript{\rm 2}University of Chinese Academy of Sciences
}


\maketitle

\begin{abstract}
   In this paper, we present a fast exemplar-based image colorization approach using color embeddings named Color2Embed. Generally, due to the difficulty of obtaining input and ground truth image pairs, it is hard to train a exemplar-based colorization model with unsupervised and unpaired training manner. Current algorithms usually strive to achieve two procedures: i) retrieving a large number of reference images with high similarity for preparing training dataset, which is inevitably time-consuming and tedious; ii) designing complicated modules to transfer the colors of the reference image to the target image, by calculating and leveraging the deep semantic correspondence between them (e.g., non-local operation), which is computationally expensive during testing. Contrary to the previous methods, we adopt a self-augmented self-reference learning scheme, where the reference image is generated by graphical transformations from the original colorful one whereby the training can be formulated in a paired manner. Second, in order to reduce the process time, our method explicitly extracts the color embeddings and exploits a progressive style feature Transformation network, which injects the color embeddings into the reconstruction of the final image. Such design is much more lightweight and intelligible, achieving appealing performance with fast processing speed.
\end{abstract}

\section{Introduction}
Image colorization aims to add vivid colors to a grayscale image for more visually pleasing appearance. As a classic vision task, image colorization has a wide range of applications, such as colorizing old photos, remastering legacy movies and transferring colors between paintings. In the past, there have produced large amounts of black-and-white photos and movies, and colorization can assist the users to give vibrant colors to these old image media, making the elapsed frozen moments relive.

In recent years, deep-learning-based colorization methods have been introduced to solve this problem. There are three types of solutions: fully automatic colorization, user-guided scribble-based colorization and exemplar-based colorization. For fully automatic colorization, existing algorithms \cite{zhang2016colorful,iizuka2016let,larsson2016learning,su2020instance} have been proposed to map the grayscale image to its color version with the supervised training. Though these methods could directly generate a colorful image without any user intervention, the colorization results are not always satisfactory.

For scribble-based colorization, it utilizes the user-selected color scribbles to guide the production of colorization. The drawback of it is that users are required with extra efforts to select abundant points in specific locations of the given grayscale image, which involves laborious and delicate work. Inappropriate scribbles and locations will lead to artifacts in the final results. Moreover, users are also required for a sense of aesthetics to choose proper and vivid colors from a palette. It is hard for an untrained user to colorize a large collection of images satisfactorily.

\begin{figure*}[t]
  \centering
  \vspace{-0.8cm}
  \includegraphics[width=0.94\linewidth]{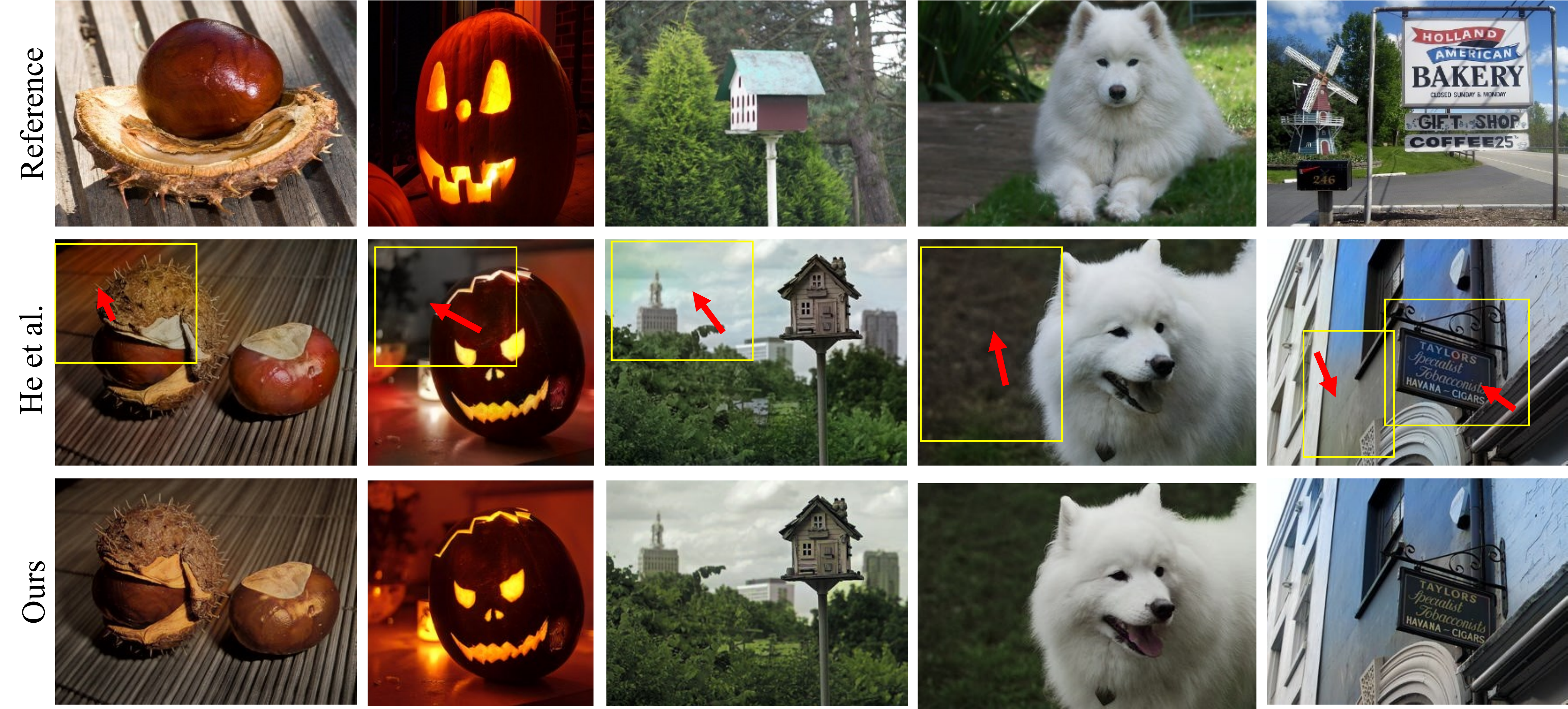}
  \caption{\textbf{First row:} reference images. \textbf{Second row:} \cite{he2018deep} could produce incorrect colors and color contaminations. \textbf{Third row:} Without explicit corresponding search, our method can avoid extra artifacts and achieve real-time processing speed with more natural visual results.}
  \label{fig:figure_motivation}
\end{figure*}

As for exemplar-based colorization, the performance is better than that of automatic colorization by adopting the additional information. However, there are two challenges for implementing exemplar-based colorization. First one is how to efficiently prepare the reference database. As the database used in image colorization is always huge (e.g., ImageNet \cite{deng2009imagenet}.), it is costly to retrieve a relevant and proper reference image for every input image. The second one is that, for an input grayscale image, there lacks its ground truth color image to construct the training pair for supervised learning. In \cite{he2018deep}, the authors use a pre-trained gray image retrieval algorithm  to construct a large reference image database. Obviously, it is extremely time-consuming and cumbersome. Moreover, if we want to conduct 
experiments on other datasets, the previous retrieval algorithm has to be retrained. As for network structure, most existing methods \cite{he2018deep,lu2020gray2colornet,iizuka2019deepremaster,zhang2019deep,Zhang_2021_WACV} explicitly compute the semantic correspondence between reference and target images with non-local operations, which will bring about extra computational expensive calculations and could lead to artifacts as shown in Figure \ref{fig:figure_motivation}.

To simplify the implementation of training such a colorization model, we use thin plate splines (TPS) \cite{duchon1977splines, chui2000new} transformation to create several self-reference images from a given color image, so that the training approach can be carried out in a paired supervised manner. Furthermore, our solution does not require any sophisticated or time-consuming processes such as the computation of correspondence maps. To achieve colorization, we extract the color information from the reference color image and then inject the color embeddings into the deep representations of the input grayscale image. Our approach, with such an efficient architecture, can achieve fast processing speed. Different from \cite{lu2020gray2colornet,zhang2019deep,lee2020reference} which adopt a bundle of loss functions, we only use the simplest pixel-wise reconstruction loss \cite{huber1992robust} and perceptual loss \cite{johnson2016perceptual} to regularize the learning procedure. These simple losses also reduce the difficulty of reproduction.

The proposed method consists of three main modules: color embeddings generation network, content embeddings generation network and Progressive Feature Formalisation Network (PFFN). The first network extracts high-dimensional feature maps from the reference image in RGB space, after which the deep feature maps are passed through a designed multi-layer perceptron (MLP) \cite{hastie2009elements} to generate the abstract embeddings. The second is devised to encode the target grayscale image into intermediate deep features. The last network takes the content and color embeddings as the input to generate a color image with the collaboration of several Progressive Feature Formalisation Blocks (PFFB), which is proposed to inject the reference color embeddings into the reconstruction process.


\begin{figure*}
  \centering
  \vspace{-0.8cm}
  \includegraphics[width=\linewidth]{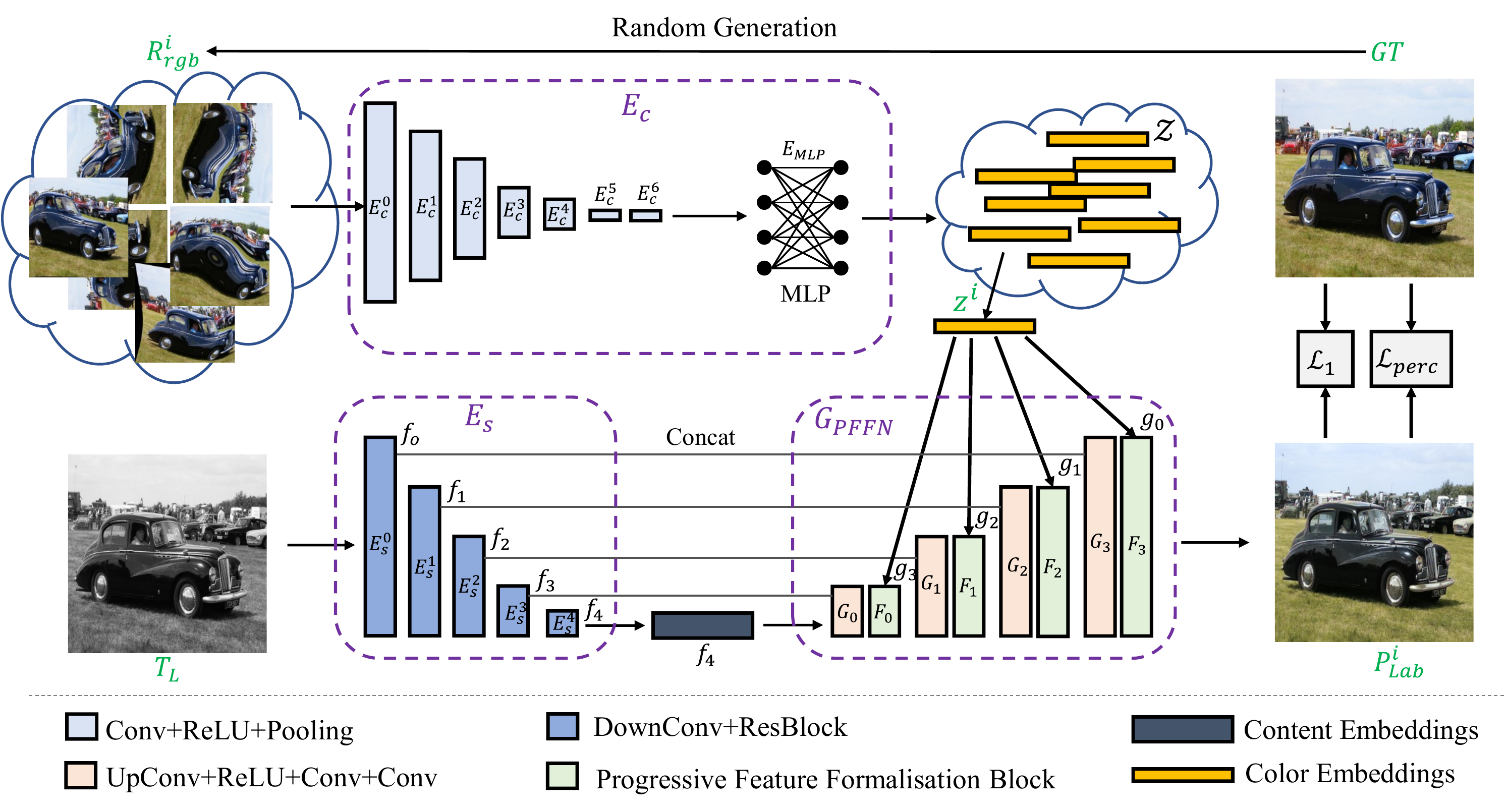}
  \caption{An illustration of the proposed Color2Embed framework, which encodes the reference images into color embeddings for exemplar-based image colorization.}
  \label{fig:figure_main}
\end{figure*}

\section{Related Work}
\subsection{Fully Automatic Colorization}
Learning-based colorization methods \cite{cheng2015deep, deshpande2015learning, larsson2016learning, zhang2016colorful,iizuka2016let,su2020instance,zhao2018pixel,zhang2017real,isola2017image,zhang2019deep,he2018deep,lu2020gray2colornet,xu2020stylization,lei2019fully,xiao2020example,Khodadadeh_2021_WACV} have addressed more attention in recent years. Rely on the strong representation ability of deep neural networks, fully automatic colorization methods \cite{cheng2015deep,deshpande2015learning, larsson2016learning, zhang2016colorful,iizuka2016let,su2020instance,zhao2018pixel} can be realized by designing various network structures and leveraging a large-scale image dataset. Iizuka et al. \cite{iizuka2016let} proposes a two-branch learning architecture to combine the global priors and local features. Zhang et al. \cite{zhang2016colorful} proposes a VGG-like deep network to output regression results and color distributions. Larsson et al. \cite{larsson2016learning} adopts the color histograms for colorization. These methods sensitively suffer from visual artifacts when the input image has complex content with multiple objects. After training, the parameters of the network would be fixed, the output colorization results are only one plausible solution without multimodality

\subsection{Scribble-based Colorization}
Using human intervention, some algorithms \cite{zhang2017real,levin2004colorization,huang2005adaptive,yatziv2006fast,qu2006manga,luan2007natural,sykora2009lazybrush} are introduced to propagate the initial color points or strokes to the entire grayscale image. The propagation is based on optimization strategy and pixel similarity metrics. In \cite{levin2004colorization}, Levin et al. adopts Markov Random Field for propagating the sparse scribble colors to the adjacent pixels which have similar luminance intensity. Qu et al. \cite{qu2006manga} and Luan et al. \cite{luan2007natural} take further consideration of image textures for propagation. Yatziv et al. \cite{yatziv2006fast} utilizes the intrinsic distance for constraint. Huang et al. \cite{huang2005adaptive} involve the edges to avoid the color blending for advanced color propagation. As for scribble-based colorization method, Zhang et al. \cite{zhang2017real} proposes a U-net-like network to interactively produce impressive results with sparse color hints and color histograms. These scribble-based colorization methods are prone to be restricted by the aesthetic of users. It is hard for an untrained user to select suitable points and select correlated colors from a palette.

\subsection{Exemplar-based Colorization}
Compared to scribble-based methods, exemplar-based colorization only requires the user to select a suitable reference image according to the target grayscale image. Earlier traditional works \cite{welsh2002transferring,reinhard2001color} transfer colors by matching the global color statistics. More early traditional methods \cite{ironi2005colorization, tai2005local,charpiat2008automatic,gupta2012image,chia2011semantic,liu2008intrinsic,bugeau2013variational} focus on adopting techniques of extracting local image features like segmented region \cite{ironi2005colorization, tai2005local, charpiat2008automatic}, super-pixel \cite{gupta2012image, chia2011semantic}, and pixel-level \cite{liu2008intrinsic, bugeau2013variational}. But these methods are vulnerable to be affected by uncorrelated regions in an image and always bring unexpected redundant outlines of objects. A pioneer work \cite{he2018deep} first proposes a deep exemplar-based colorization algorithm that utilizes the pre-trained VGG-19 to extract deep features for similarity calculation and then warps reference image. Xiao et al. \cite{xiao2020example} proposes a pyramid structure network to exploit the inherent multi-scale color representations. Lu et al. \cite{lu2020gray2colornet} proposes an approach that jointly considers the semantic correspondence and global color histograms into network design. Xu et al. \cite{xu2020stylization} proposes a fast deep exemplar-based method by adopting an AdaIN-based \cite{huang2017arbitrary} transfer network and randomly generates color hints of the transferred image. Such random sampling procedure prone to produce unstable colors by the incorrect selection of color hints.

\section{Method}

\begin{figure*}[htp]
  \centering
  \vspace{-0.8cm}
  \includegraphics[width=0.9\linewidth]{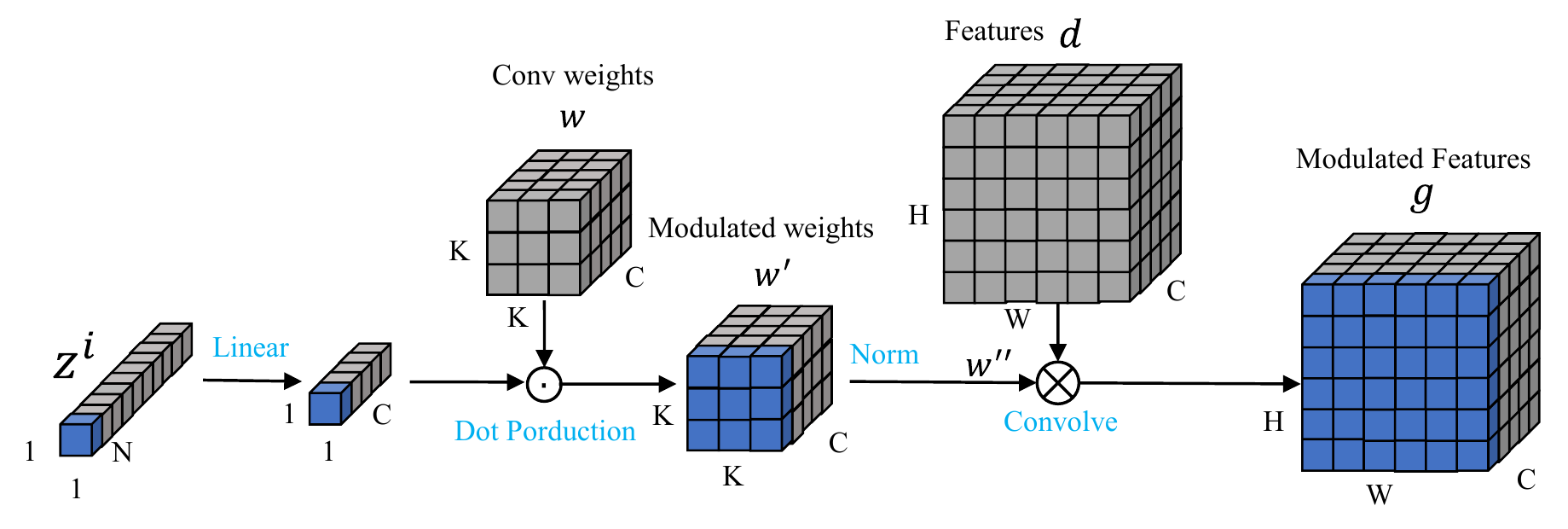}
  \caption{The illustration of progressive feature formalisation block.}
  \label{fig:figure_PFFB}
\end{figure*}

\subsection{Overview}
Following previous methods \cite{zhang2016colorful, zhang2017real, zhang2019deep, he2018deep, lu2020gray2colornet, xu2020stylization, xiao2020example, iizuka2016let}, we perform this task in CIE Lab \cite{schanda2007cie} color space. Each image can be separated into a luminance channel $L$ and two chrominance channels $a$ and $b$. Given an $H \times W$ grayscale target image $T_L \in R^{H \times W \times 1}$ containing only the luminance channel and a color reference image $R_{rgb} \in R^{H\times W \times 3}$ which is represented in RGB space, the aim of exemplar-based colorization is to find a function $F(T_L|R_{rgb})$ that predicts the corresponding $a$ and $b$ channels $P_{ab}$. Different from previous methods, we directly extract the color embeddings in RGB color space while our reconstruction process is performed in Lab color space. 

The overall structure of the proposed end-to-end network is shown in Figure \ref{fig:figure_main}. $E_c$, $E_s$ and $G_{PFFN}$ are color encoder, content encoder and Progressive Feature Formalisation Networks (PFFN) respectively. $E_c$ takes the random reference $R_{rgb}^{i}$ as input and generates the color embeddings $z^i$. We denote $Z$ as the set of all color embeddings. 
\begin{equation}
\begin{aligned}
z^{i} = E_c(R_{rgb}^{i}).
\end{aligned}
\end{equation}
$E_s$ represents the content encoder network which contains several downsampling convolutions and resblocks to extract the intermediate features and content embbedings. 
\begin{equation}
\begin{aligned}
f_N = E_s^{N-1}(f_{N-1}).
\end{aligned}
\end{equation}
These intermediate features are passed to $G_{PFFN}$ for providing multi-scale content information. $G_{PFFN}$ contains consecutive upsampling blocks and Progressive Feature Formalisation Blocks (PFFB).
\begin{equation}
g_i = \left\{
\begin{aligned}
    & F_i(G_i(f_N, f_{N-1})), if \ i=N-1, \\
    & F_i(G_i(g_{i+1}, f_{i})), otherwise,
\end{aligned}
\right.
\end{equation}
where $i \in \{ 1, ..., N-1 \}$, $G_i$ represents the upsampling module and $F_{i}$ represents the PFFB.
For easily implementation, we only adopt two wildly used loss functions, reconstruction loss and perceptual loss. Remarkably, during training, we generate $R_{rgb}$ from ground truth by applying the Thin Plate Splines (TPS) transformation \cite{duchon1977splines, chui2000new}. During testing, our model can accept any given color images as references to colorize the grayscale image.

In the following, we will introduce Self-Augmented Self-Reference Learning in Sec. 3.2, PFFN in Sec. 3.3, and loss functions in Sec. 3.4.

\subsection{Self-Augmented Self-Reference Learning}
As preparing a reference image for a grayscale image and conjugating these two inputs for pixel-wise paired training is an obstacle, we adopt the reference generation method from \cite{lee2020reference}. To generate random reference image $R_{rgb}^{i}$ for a given target image $T_L$, we apply spatial transformations on ground truth image $GT$. Since $R_{rgb}^{i}$ is essentially generated from $GT$, these processes guarantee the sufficient color information to colorize $T_L$, which encourage the proposed framework to reflect the $R_{rgb}^{i}$ in the colorization process. The detail on how these transformations operate is described in the following. First, the content transformation $C(\cdot)$ adds a particular random noise on $GT$. The reason why we impose the noise of the original $GT$ is to increase the training samples. The same ground truth $GT$ could have various reference images. Afterwards, we further apply TPS transformation $T(\cdot)$, a non-linear spatial transformation on $C(GT)$, resulting in our final reference image $R_{rgb}^{i}$. This prevents our model from lazily bringing the color in the same pixel location from $GT$ while enforcing the color Encoder $E_c$ to only extract the semantic color information from a reference image even with a spatially different layout. The above two transformations help our model learn to transfer color information from an exemplar image to a target image and avoid learning an identity mapping.


\textbf{Discussion.} In the proposed method, we first encode the reference image into a lower-dimensional embeddings which is supposed to represent the color information. Why does such mechanism work? How to ensure that the extracted embeddings contain only color information without image content information? As shown in Figure \ref{fig:figure_main}, during training, given an original reference color image $GT$, we can obtain multiple reference versions {$R_{rgb}$} with different geometric transformations, which are not aligned in image content but contain coherent color characteristics. More importantly, these generated reference images are not aligned with the target image $T_L$ in pixel-level as well. They are demanded to only provide the color guidance for $T_L$. Thus, the supervision signal only acts on the reconstruction of color without consideration of the image content. Extensive experiments have also demonstrated the effectiveness of injecting such color embeddings for obtaining exemplar-based colorization results.

\subsection{Progressive Feature Formalisation Network}
To formalize the content features, we embed the feature modulation module from StyleGAN2 \cite{karras2020analyzing} into our PFFN. As shown in Figure \ref{fig:figure_PFFB}, the convolution weights $w$ is the initial trainable weights, and we first scale it with a constant hyperparameter $s$ corresponding to the channel dimension of the input feature maps and project it with a linear operation to align the channel size. It can be formulated by
\begin{equation}
\begin{aligned}
w^{'} = w\cdot s \cdot F_{Linear}(z),
\end{aligned}
\end{equation}
where $w$ and $w^{'} \in R^{C_i \times C_j \times K \times K}$ are the original and modulated convolution weights, respectively. The kernel size of convolution weights is $K$ while $C_i$ and $C_j$ represent the number of input and output channels. The $F_{Linear}$ transfers the color embeddings $z^i$ to modulate the weights $w$ in this feature scale. After the modulation, we adopt the normalization procedure with $F_{Norm}$ to further constrain the values of convolution weights. The $F_{Norm}$ can be formulated as
\begin{equation}
\begin{aligned}
F_{Norm} = \frac{1}{\sqrt{\sum {w^{'}}^{2} + \epsilon}},
\end{aligned}
\end{equation}
where the $\epsilon$ is a small constant to avoid the numerical issues. As for the detail of $F_{Norm}$, we normalize the dimension of $w^{'}$ except from the input feature dimension. This operation can be regarded as a kind of vector unitization, making the convolution weights focus more on the direction rather than the absolute values. From our experience, this operation also stables the output performance during training. As depicted in Figure \ref{fig:figure_PFFB}, $F_{Linear}$ and $F_{Norm}$ represent ("Linear") and (operations, respectively. The final convolution weights is $w^{''}$, which can be formally written by
\begin{equation}
\begin{aligned}
w^{''} = F_{Norm}(w^{'}).
\end{aligned}
\end{equation}
Given a content features $d$, the modulated features $g$ can be formulated by
\begin{equation}
\begin{aligned}
g = F_{convolve}(w^{''}, d),
\end{aligned}
\end{equation}
where $F_{convolve}$ represents convolution operation.

\subsection{Loss Function}
Most existing methods tend to devise and utilize a variety of complex loss functions to achieve multiple constrains in their design. Ascribing to the effective and ingenious design of our method, we only need to adopt two widely-used loss functions to accomplish satisfactory colorization performance.

\textbf{Reconstruction Loss.}
Owing to the self-augmented self-reference training mechanism, it is able to conduct paired supervised learning, which can make the training process more stable and accelerate the convergence. We adopt smooth $L1$ loss \cite{huber1992robust} as the distance metric to avoid the averaging solution in the ambiguous colorization problem \cite{zhang2017real}. The reconstruction loss can be formulated as:
\begin{equation}
\begin{aligned}
\mathcal{L}_{recon} = \sum_{x} smooth {L1}(P_{ab}^{i}(x), GT_{ab}(x)),
\end{aligned}
\end{equation}
where $smooth {L1}(a, b) = \frac{1}{2} (a-b)^2, if |a-b|< 1, smooth_{L1}(a,b) =|a-b| - \frac{1}{2}, otherwise$.

\textbf{Perceptual Loss.}
To allow the network predict the perceptually plausible colors even without a proper reference image, we adopt a perceptual loss \cite{johnson2016perceptual} to constrain the predicted $P_{rgb}^{i}$ and $GT$. The $P_{rgb}^{i}$ is transformed from $P_{Lab}^{i}$ with a color space transformation and $P_{lab}^{i}$ is concatenated from input grayscale image $T_L$ and output $P_{ab}^{i}$. Formally,
\begin{equation}
\begin{aligned}
\mathcal{L}_{perc}=\sum_{x}||F_P(x) - F_{GT}(x)||_1,
\end{aligned}
\end{equation}
where $F_P$ and $F_{GT}$ represent the feature maps extracted from VGG-19 \textit{relu5\_2} layer from $P_{rgb}^{i}$ and $GT$, respectively.

In summary, the overall loss function for training is defined as:
\begin{equation}
\begin{aligned}
\mathcal{L}_{total}=\lambda_{rec} \mathcal{L}_{recon} + \lambda_{perc} \mathcal{L}_{perc},
\end{aligned}
\end{equation}
where the $\lambda_{rec}$ and $\lambda_{perc}$ represent the weights of the two losses, respectively.

\begin{figure}[htb]
  \centering
  \includegraphics[width=\linewidth]{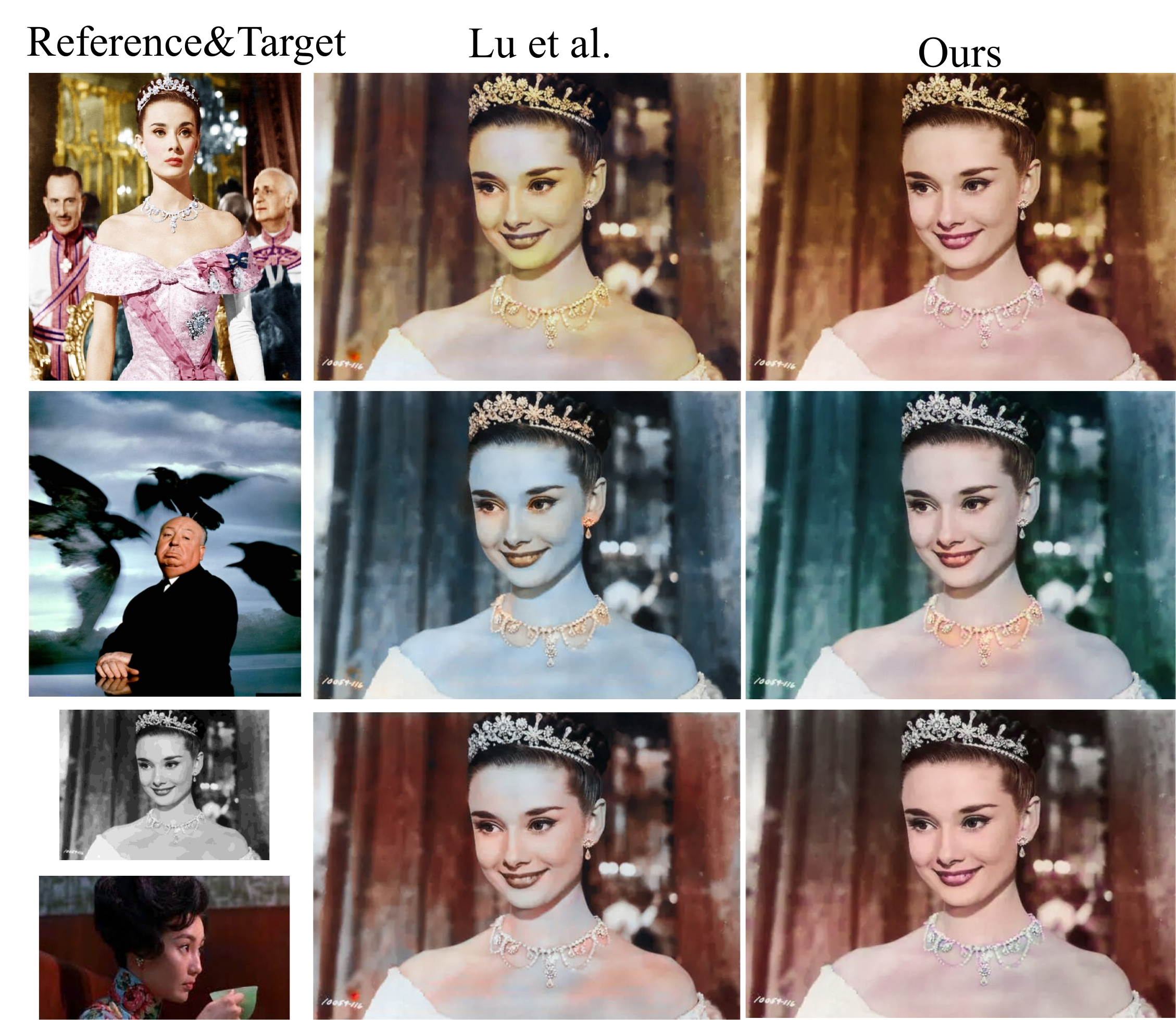}
  \caption{Comparison with Lu et al. \cite{lu2020gray2colornet} on Image \textit{Hepbum}.}
  \label{fig:figure_compare_hepbum}
\end{figure}

\begin{figure*}[ht]
  \centering
  \vspace{-0.8cm}
  \includegraphics[width=0.95\linewidth]{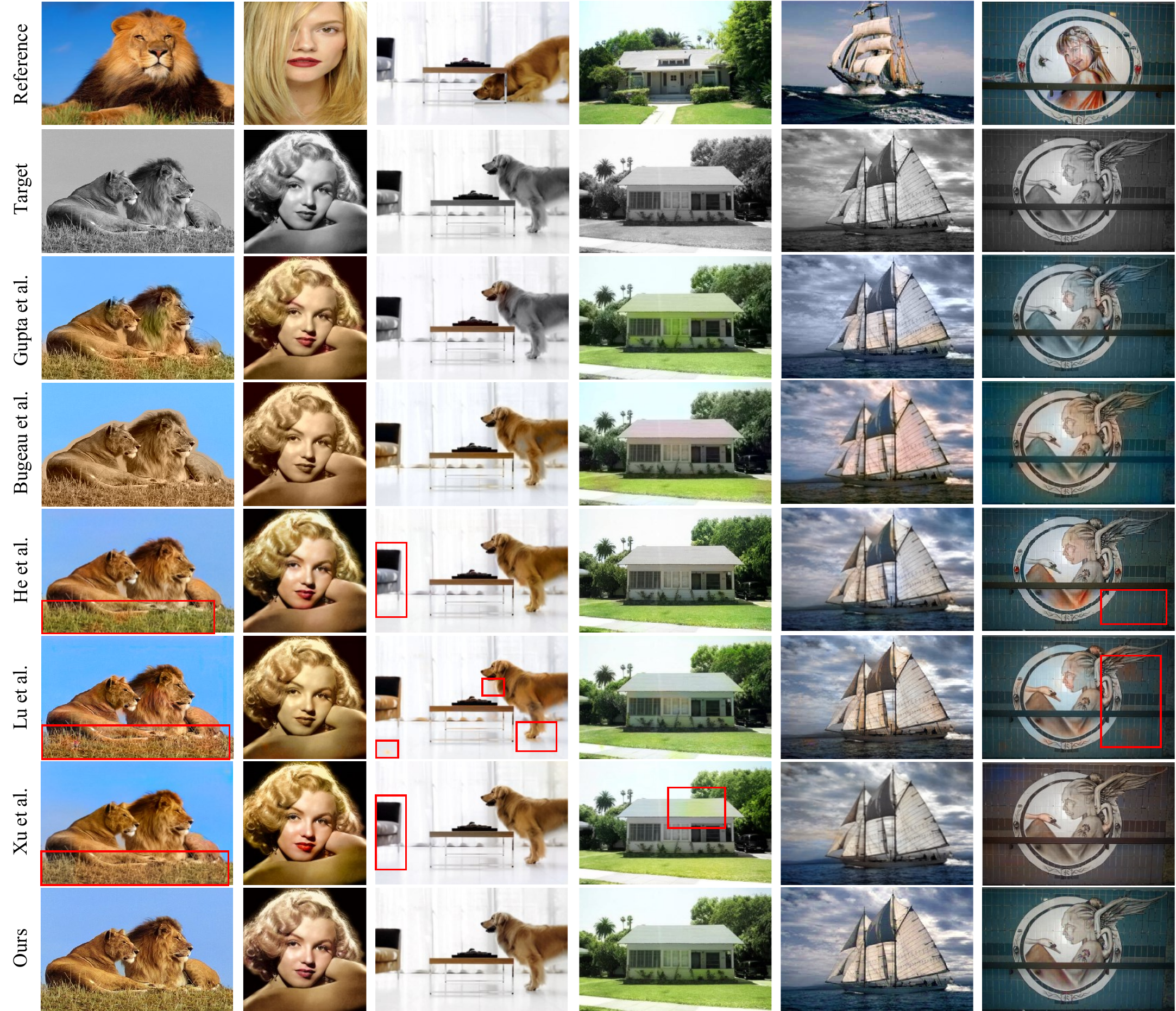}
  \caption{Comparison with state-of-the-art exemplar-based colorization methods. The first two rows are input target-reference image pairs. The last six rows are corresponding colorization results generate by \cite{bugeau2013variational, gupta2012image, he2016deep, lu2020gray2colornet, xu2020stylization} and the proposed method, respectively. \textcolor[RGB]{246,0,0}{Red rectangles} highlight failures and artifacts. Please zoom in for best view.} 
  \label{fig:figure_comparison}
\end{figure*}

\section{Experiments}
\subsection{Implementation Detail}
\textbf{Dataset and Metric.}
ImageNet\cite{deng2009imagenet} is a commonly used image classification dataset with 1.2 million images from 1000 categories. Following previous image colorization methods \cite{zhang2016colorful, zhang2017real, he2018deep, lu2020gray2colornet}, we adopt this dataset as our ground truth color image dataset to generate reference images and grayscale images.  We resize all training images into size $256 \times 256$ with \textit{bilinear} rescaling method. During generating reference image, we add random Gaussian noise with mean $0$ and variance $\sigma = 5$. The geometric transformation of TPS is randomly applied, hence, the same image has various transformed pairs every time. For data augmentation, we randomly rotate and filp the reference image. As for test datasets, we use the images from He et al. \cite{he2018deep} for a fair comparison.

\begin{figure*}[htb]
  \centering
  \includegraphics[width=0.95\linewidth]{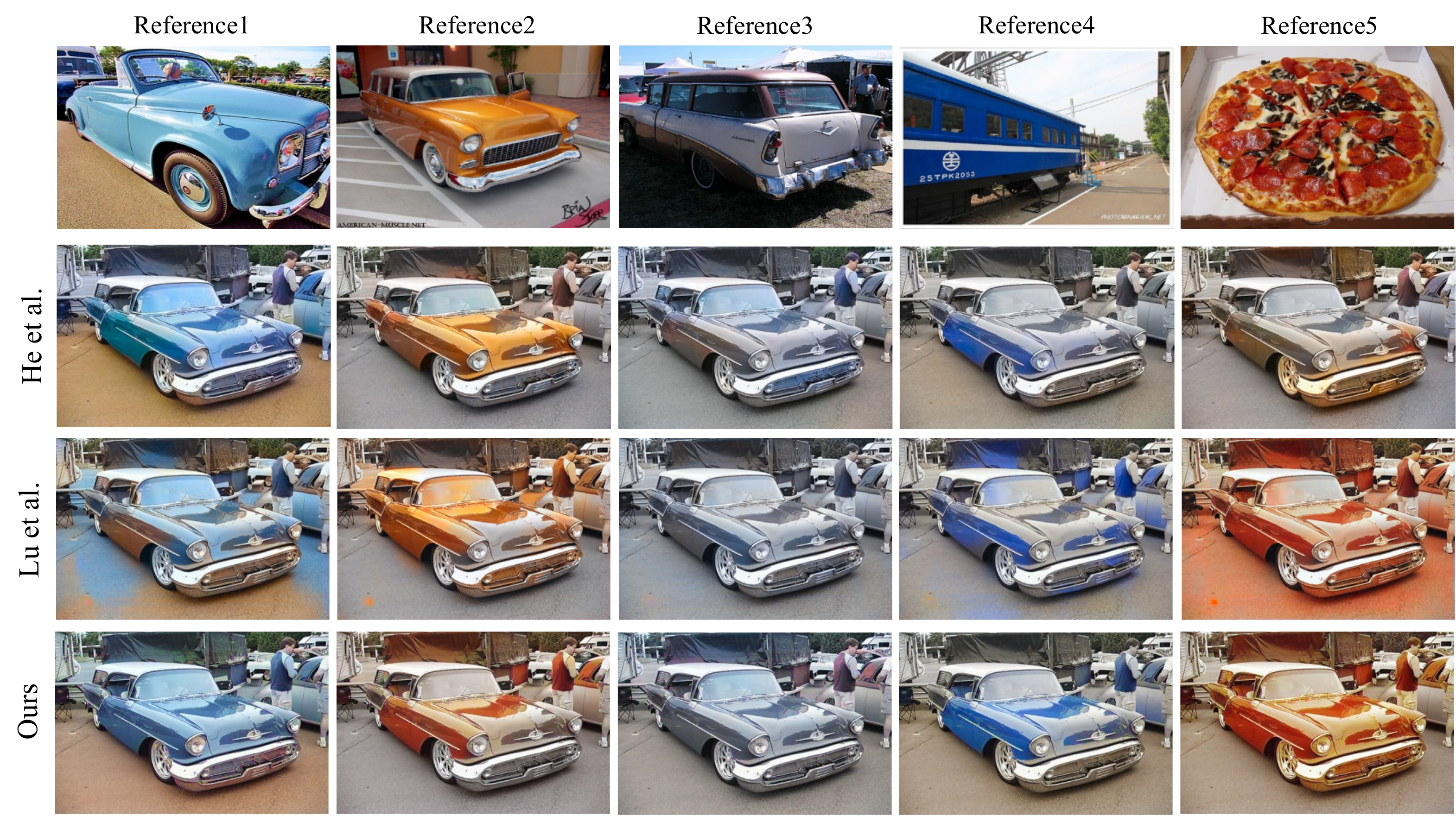}
  \caption{Results of colorization on different reference images. When the reference image content is not similar with target image, He et al. \cite{he2018deep} and Lu et al. \cite{lu2020gray2colornet} produce more artifacts due to its explicit computation of correspondence map.}
  \label{fig:figure_multi_reference}
\end{figure*}

\begin{figure*}[htb]
  \centering
  \includegraphics[width=0.92\linewidth]{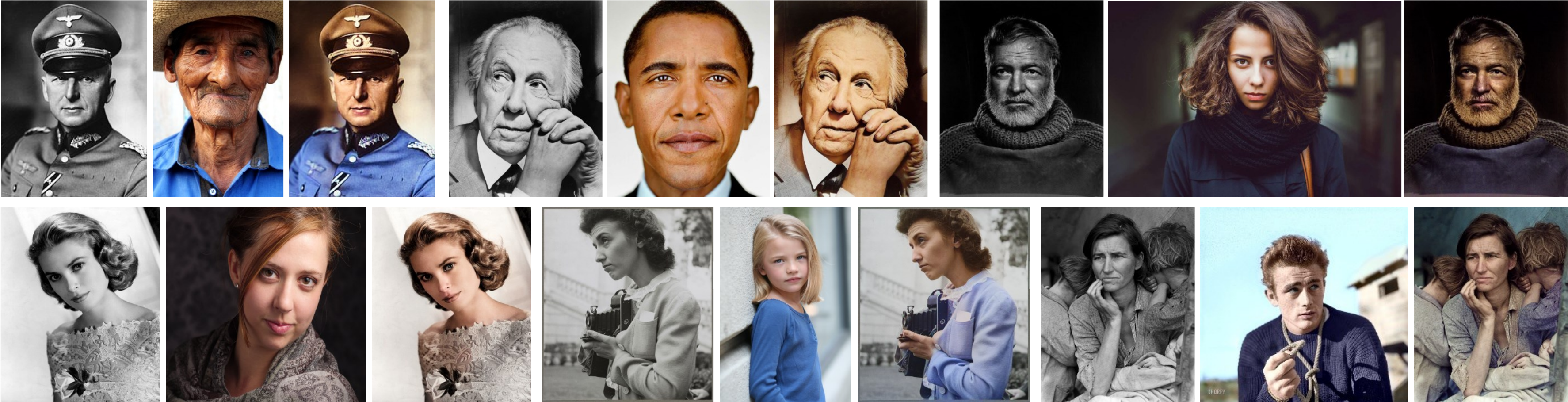}
  \caption{Results of legacy pictures. In each set, the target image, the reference and ours result lie on the left, middle, and right, respectively.}
  \label{fig:figure_human}
\end{figure*}

\textbf{Training Details.}
The $H$ and $W$ is 256, $C$ are both 512. The size of color embeddings $z^{i}$ is $512 \times 1 \times 1$. When training, we use PyTorch \cite{NEURIPS2019_9015} framework to train our model with batch size 64 on 8 NVIDIA 1080Ti GPUs. The learning rate is $1 \times 10^{-4}$ using the Adam \cite{kingma2014adam} optimizer with parameters $\beta_1=0.9$ and $\beta_2=0.99$. We set $\epsilon = 1 \times 10^{-8}, \lambda_{rec}=1, \lambda_{perc}=0.1$.

\subsection{Comparison with State-of-the-art Methods}
To evaluate the effectiveness of our method, we compare our method with five state-of-the-art methods: Gupta et al. \cite{gupta2012image}, Bugeau et al. \cite{bugeau2013variational}, He et al. \cite{he2018deep}, Lu et al. \cite{lu2020gray2colornet}, Xu et al. \cite{xu2020stylization}. These methods are all exemplar-based colorization approaches, where \cite{gupta2012image} and \cite{bugeau2013variational} are traditional methods, \cite{he2018deep, lu2020gray2colornet, xu2020stylization} are deep learning-based methods. To provide a fair comparison, we directly borrow their released results or run their test code to generate results.


\textbf{Qualitative Evaluation.}
To compare with the existing exemplar-based methods, we run our algorithm on 35 pairs collected from \cite{he2018deep}. Figure \ref{fig:figure_comparison} shows some representatives. The traditional methods \cite{gupta2012image} and \cite{bugeau2013variational} bring about many artifacts in the first image. \cite{gupta2012image} leads to unbalanced colors in all five images and \cite{bugeau2013variational} has unexpected colors or redundant outlines of objects. \cite{he2018deep} tends to bring color contamination at image local regions like in the first, second, fourth, and fifth images. \cite{lu2020gray2colornet} is a newly proposed method that first incorporates the local semantic correspondence and global color histogram for consideration to generate the colorized image. In Figure \ref{fig:figure_comparison}, due to the unstable influence of semantic correspondence, it also brings about color contamination in the first and fifth images. To reveal the advantage of our approach, we compare with two methods with explicit computation of correspondence map in Figure \ref{fig:figure_motivation} and Figure \ref{fig:figure_compare_hepbum}.

\textbf{Robustness of diverse references.}
In order to evaluate the robustness of diverse references, we additionally provide the qualitative comparison on five reference images in Figure \ref{fig:figure_multi_reference}. As mentioned in Figure \ref{fig:figure_motivation}, \cite{he2018deep} and \cite{lu2020gray2colornet} tend to bring unexpected artifacts due to the incorrect computation of correspondence map.

\begin{figure}
  \centering
  \includegraphics[width=\linewidth]{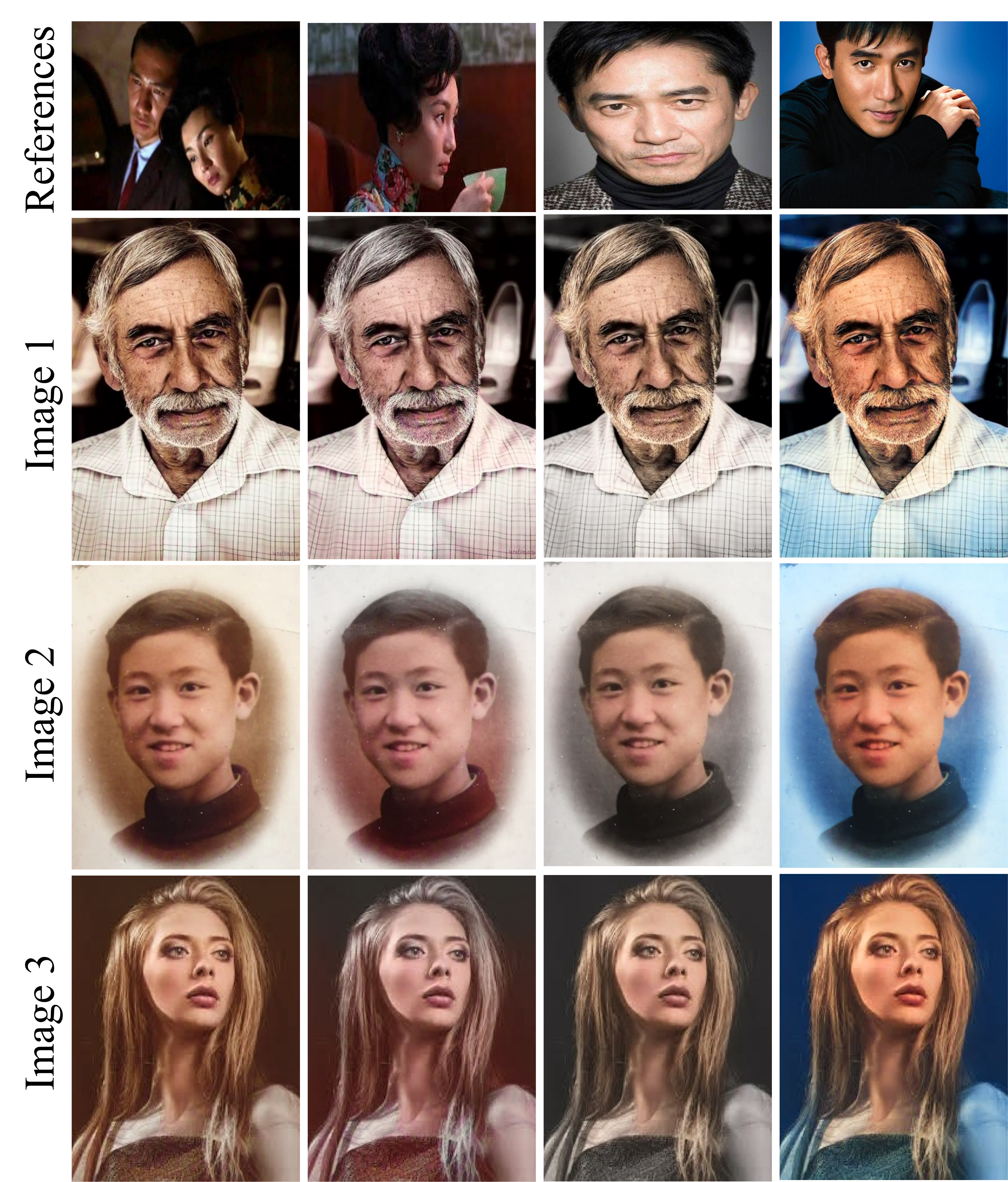}
  \caption{Different reference images will change the color style in the results. Our method has the robustness of accepting diverse references.}
  \label{fig:figure_film}
\end{figure}

\textbf{Results on old photos.}
We further evaluate the performance of Color2Embed on real old photos collected from the internet, as shown in Figure \ref{fig:figure_human} and Figure \ref{fig:figure_film}.

\linespread{1.1}
\begin{table}[htb]
\caption{\label{tab:running-name}The running time (second) comparison of exemplar-based colorization methods. OOM denotes out of memory. All the experiments are tested on single NVIDIA 1080Ti GPU.}
\vspace{8pt}
\setlength{\tabcolsep}{0.4mm}{
\begin{tabular}{lcccl}
\cline{1-4}
Image Size  & 256$\times$256     & 512$\times$512     & 1024$\times$1024   &  \\ \cline{1-4}
He et al. \cite{he2018deep}   & 7.25+1.14s   & 33.69+1.30s  & 49.96+OOM   &  \\
Xiao et al. \cite{xiao2020example} & 0.48s        & 1.4s         & 3.96s        &  \\
Lu et al. \cite{lu2020gray2colornet}   & 0.44s        & 1.37s        & OOM         &  \\
Xu et al. \cite{xu2020stylization}  & 0.04+0.08$\times$2s        & 0.14+0.12$\times$2s        & 0.59+0.28$\times$2s        &  \\
Ours        & \textbf{0.03}s        & \textbf{0.08}s        & \textbf{0.32}s        &  \\ \cline{1-4}
\end{tabular}%
}
\end{table}

\textbf{Running Time.} In this study, we focus on providing a fast algorithm by removing the computation of correspondence map. We compare with other four exemplar-based colorization methods to illustrate our advantage over running time in Table \ref{tab:running-name}. Notice that He et al. \cite{he2018deep} and Xu et al. \cite{xu2020stylization} have two stages: color transfer stage and colorization stage. They cost much more time than others.

\begin{figure}[htb]
  \centering
  \includegraphics[width=\linewidth]{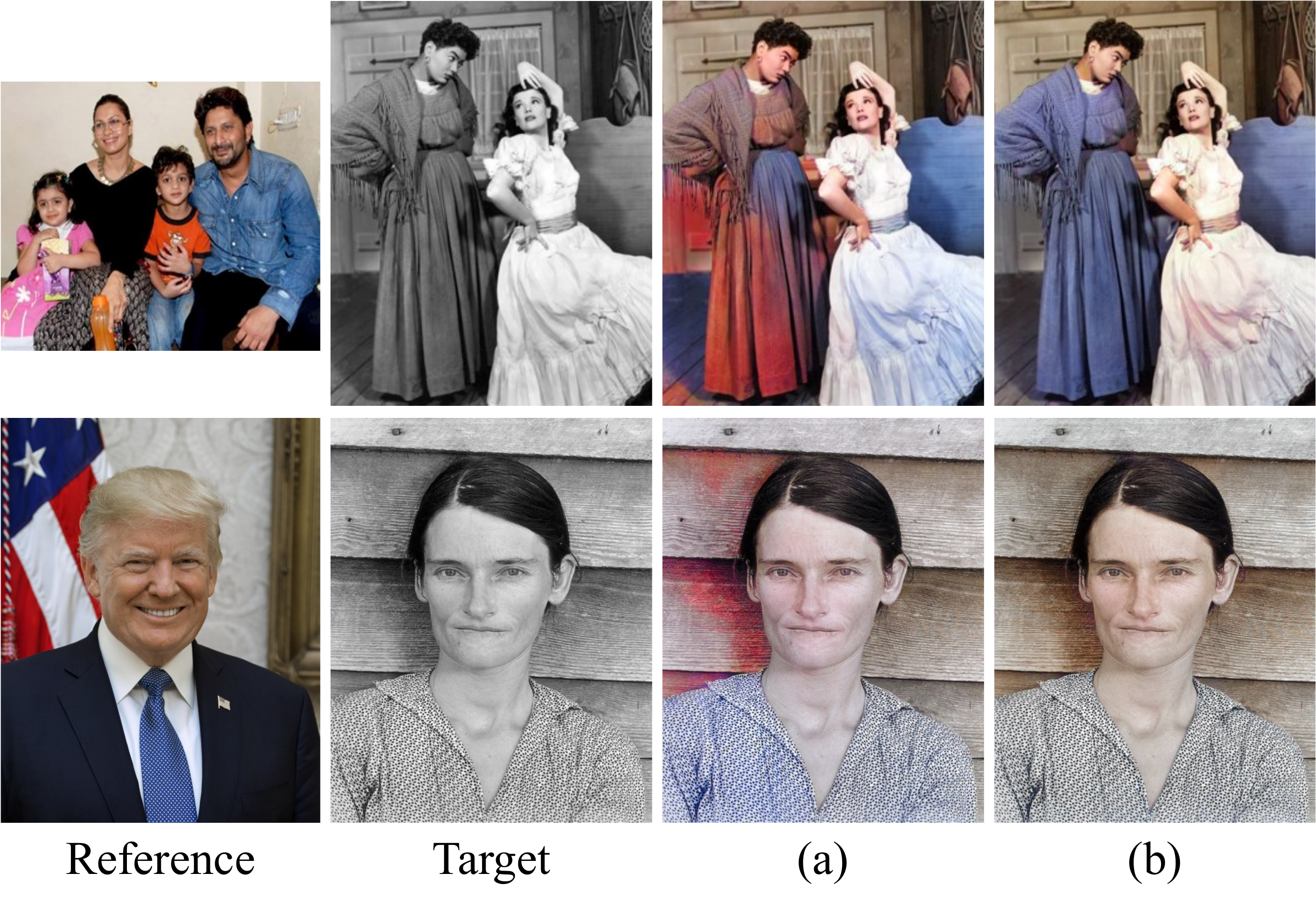}
  \caption{Effectiveness of self-augmented self-reference learning. (a) without self-augmentation. (b) with self-augmentation.}
  \label{fig:figure_ablation_selfaug}
\end{figure}

\section{Ablation Study}

\textbf{Effectiveness of Self-Augmented Self-Reference Learning.}
To evaluate the effectiveness of self-augmented self-reference learning, we train Color2Embed directly with the ground truth as reference without self-augmented operation. Figure \ref{fig:figure_ablation_selfaug} (a) appears obvious color contamination since the model learns to retain positions of these colors in the reference image. On the contrary, Figure \ref{fig:figure_ablation_selfaug} (b) successfully extracts colors from reference and then propagates them to appropriate regions.

\begin{figure}
    \centering
    \subfigure[Four examples for evaluating the effectiveness of perceptual loss.]{
    \includegraphics[width=\linewidth]{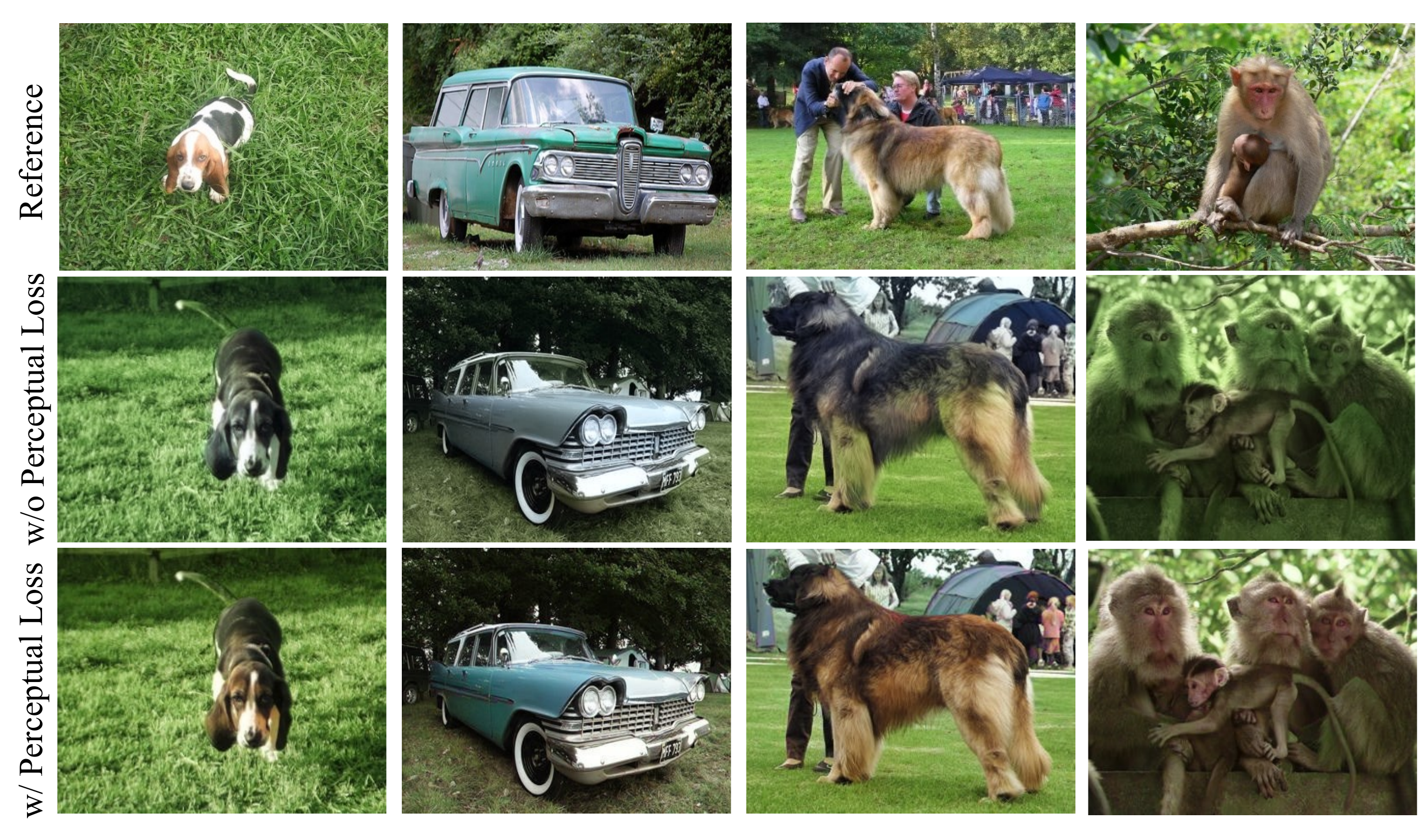}
    \label{fig:figure_ablation_onlyL1}
    }
    \subfigure[The effectiveness of reconstruction loss. 
    ]{
    \includegraphics[width=\linewidth]{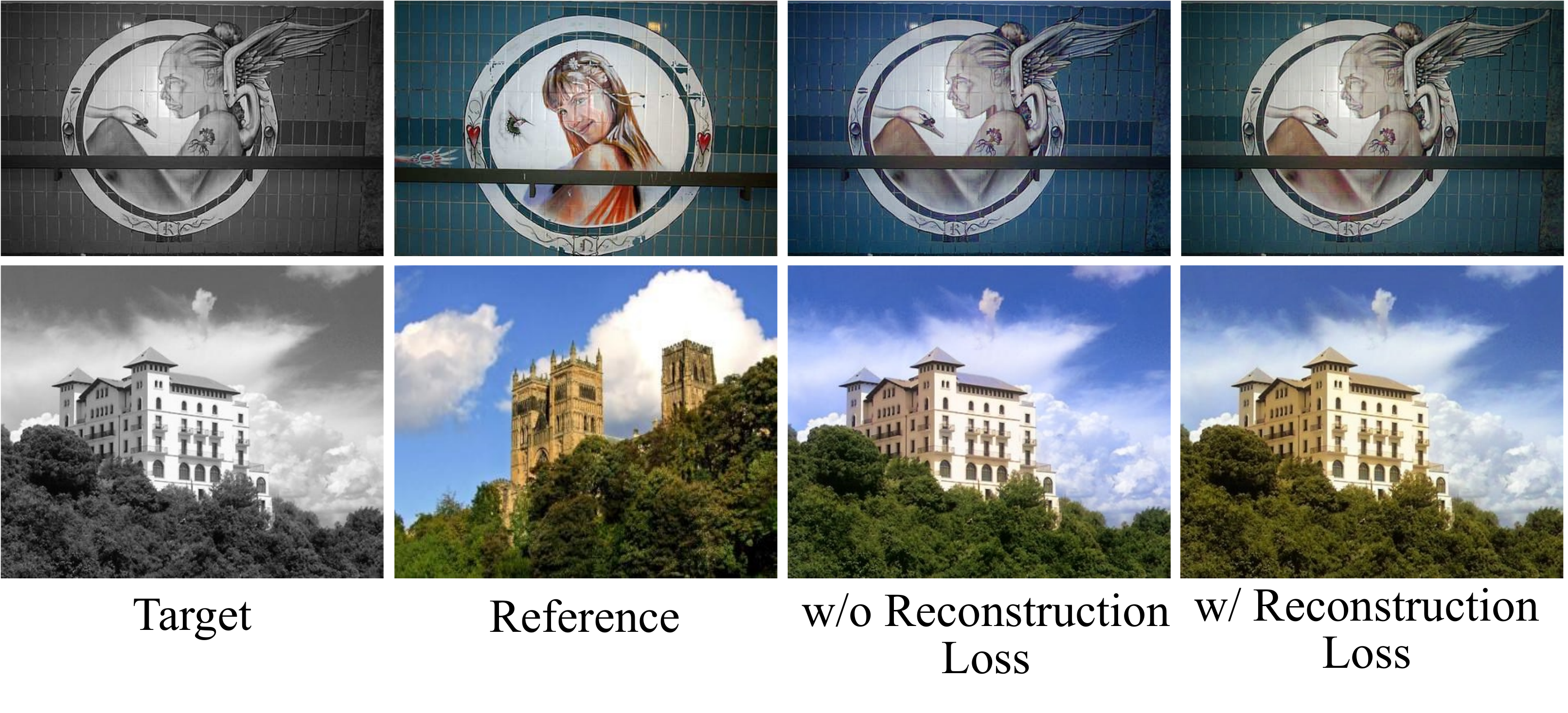}
    \label{fig:figure_ablation_onlyPerc}
    }
    \caption{Effectiveness of loss functions. Only one single loss leads to unsatisfactory performance and accurate colors are observed by combining these two losses.}
\end{figure}

\textbf{Effectiveness of Loss functions.}
To evaluate the effectiveness of loss functions, we train Color2Embed with single reconstruction or perceptual loss, respectively in Figure \ref{fig:figure_ablation_onlyL1} and \ref{fig:figure_ablation_onlyPerc}. When train Color2Embed without perceptual loss, the model will produce unsatisfactory results as shown in Figure \ref{fig:figure_ablation_onlyL1}. The second row of dog, car and monkey images obtain incorrect colors. We demonstrate the effectiveness of reconstruction loss will influence the coherence between target and reference images. In the first row of Figure \ref{fig:figure_ablation_onlyPerc}, the third image produce unseen colors of background which is really different with reference. In the second row, reference image shows a yellow tower, while the building of third image shows white color.




\section{Conclusion}
In this paper, we present a fast exemplar-based colorization algorithm named Color2Embed. Our method first converts the reference color images into color embeddings and then utilize the extracted color embeddings to generate robust results by the proposed progressive feature formalisation network. The variety of training pairs ensure the color embeddings generation network only extract color information from reference. This mechanism makes our approach have the ability of receiving any color images as reference without bringing severe artifacts. Furthermore, we typically find that our method does no need to adopt excess loss functions compared to existing methods. Extensive experiments show that our results surpass other state-of-art methods in qualitative comparison and running time.

{\small
\bibliographystyle{ieee_fullname}
\bibliography{color2embedding}
}

\end{document}